\documentclass{article}
\usepackage{spconf,amsmath,graphicx}
\input{mysymbol.sty}

\usepackage{siunitx} 
\usepackage{multicol,footnote}
\usepackage{xcolor}
\usepackage{graphicx}
\colorlet{RED}{red}
\usepackage{booktabs}
\usepackage{algorithm2e}
\RestyleAlgo{ruled}
\usepackage{algpseudocode}
\usepackage{longtable}
\usepackage{relsize}
\usepackage{makecell}
\usepackage{array}
\usepackage[export]{adjustbox}
\usepackage{lipsum}
\usepackage{amsmath,nccmath}
\usepackage{multirow}
\usepackage{subcaption}
\usepackage{todonotes} 
\usepackage{amssymb}

\usepackage{amsmath,amsfonts,bm}

\def\eqref#1{equation~\ref{#1}}

\def\1{\bm{1}}

\DeclareMathAlphabet{\mathsfit}{\encodingdefault}{\sfdefault}{m}{sl}
\SetMathAlphabet{\mathsfit}{bold}{\encodingdefault}{\sfdefault}{bx}{n}

\usepackage{mathtools}
\usepackage{multicol}
\usepackage{booktabs}
\DeclarePairedDelimiterX{\norm}[1]{\lVert}{\rVert}{#1}
\usepackage{hyperref}
\usepackage{url}
\usepackage{multicol}
\usepackage{multirow}
\usepackage{enumitem}
\usepackage{xcolor}
\usepackage{algorithm2e}
\RestyleAlgo{ruled}
\usepackage{algpseudocode}
 
\usepackage{amssymb,amsmath,amsthm, amsfonts}
\newtheorem{theorem}{Theorem}

\newtheorem{proposition}{Proposition}
\definecolor{morange}{rgb}{0.8,0.2,0}
\definecolor{mblue}{rgb}{0,0.3,1.0}
\definecolor{mred}{rgb}{0.9,0.1,0.1}
\definecolor{mpurple}{rgb}{0 0 0}

 \newcommand{\ignore}[1]{}
\theoremstyle{plain}

\theoremstyle{definition}
\theoremstyle{remark}
\newtheorem{remark}[theorem]{Remark}

\long\def\symbolfootnote[#1]#2{\begingroup
 	\def\thefootnote{\fnsymbol{footnote}}
 	\footnote[#1]{#2}\endgroup}
\allowdisplaybreaks[4]

\title{ Fairness-Aware Graph Filter Design }
\name{O. Deniz Kose$^{\star}$ \qquad Yanning Shen$^{{\star}}$ \qquad Gonzalo Mateos$^{\dagger}$}
  
  \address{$^{\star}$ Department of Electrical Engineering and Computer Science, University of California Irvine, USA\\
 $^{\dagger}$ Department of Electrical and Computer Engineering, University of Rochester, USA}

\begin{document}



\maketitle

\let\thefootnote\relax\footnotetext{Work in this paper was supported in part by the Google Research Scholar Award, and the NSF awards CCF-1750428 and CCF-1934962.} 


\begin{abstract}
Graphs are mathematical tools that can be used to represent complex real-world systems, such as financial markets and social networks. Hence, machine learning (ML) over graphs has attracted significant attention recently. However, it has been demonstrated that ML over graphs amplifies the already existing bias towards certain under-represented groups in various decision-making problems {\color{mpurple} due to the information aggregation over biased graph structures}. Faced with this challenge, in this paper, we design a fair graph filter that can be employed in a versatile manner for graph-based learning tasks. The design of the proposed filter is based on a bias analysis and its optimality in mitigating bias compared to its fairness-agnostic counterpart is established. Experiments on real-world networks for node classification demonstrate the efficacy of the proposed filter design in mitigating bias, while attaining similar utility and better stability compared to baseline algorithms. 
\end{abstract}
%


\section{Introduction}

We live in the era of connectivity, where the actions of humans and devices are increasingly driven by their relations to others. Concurrently, a significant amount of data that describes different interconnected systems, such as social networks, Internet of Things (IoT), the Web, and financial markets, is increasingly available. Processing and learning from such data can provide significant understanding and advancements for the corresponding networked systems. In this context, machine learning (ML) over graphs has attracted increasing attention \cite{chami2022machine, gcn}, since graphs are widely utilized to represent complex underlying relations in real-world networks~\cite{mateos19spmag}. 

These relational patterns can be captured by graph edges, while attributes of nodes (nodal features) can be interpreted as signals defined on the vertices. For example, in a social network, user ages can be modeled as a graph signal, and the friendship information can be captured by the edges. Graph signal processing (GSP) \cite{gsp} extends the tools in classical signal processing to graph signals, such as frequency analysis, sampling and filtering \cite{gft, dgsp,  marques2017stationary, isufi2016autoregressive}. GSP and ML over graphs are closely intertwined, where the tools in one domain can be useful in the other one \cite{gsp, dong2020graph}. For instance, it has been demonstrated that graph neural networks (GNNs) can be designed, analyzed, and improved by leveraging GSP-based insights \cite{gama2020graphs, gama2020stability, dong2020graph}, which underscores the advancements that can be made  by cross-pollinating the findings in both domains. 

\noindent \textbf{Fairness in ML.} Despite the growing interest towards learning over graphs, the widespread deployment of these algorithms in real-world decision systems depends heavily on how socially responsible they are. Indeed, several studies have demonstrated that ML models propagate the historical bias within the training data and lead to discriminatory results in ensuing applications \cite{ beutel2017data}. Specific to graph-based learning, the utilization of graph structure in the algorithm design has been shown to amplify the already existing bias \cite{fairgnn}. Motivated by this, recent works focus on fairness-aware learning over graphs and advocate different techniques, such as adversarial regularization \cite{bose2019compositional, fairgnn}, fairness constraints \cite{buyl2021kl, kose2022fairnorm}, and fairness-aware graph data augmentation \cite{kose2022fair, spinelli2021fairdrop, dong2022edits}. 
\begin{figure*}[t]
    \centering
    \begin{subfigure}{0.35\textwidth}
        \includegraphics[width=\textwidth]{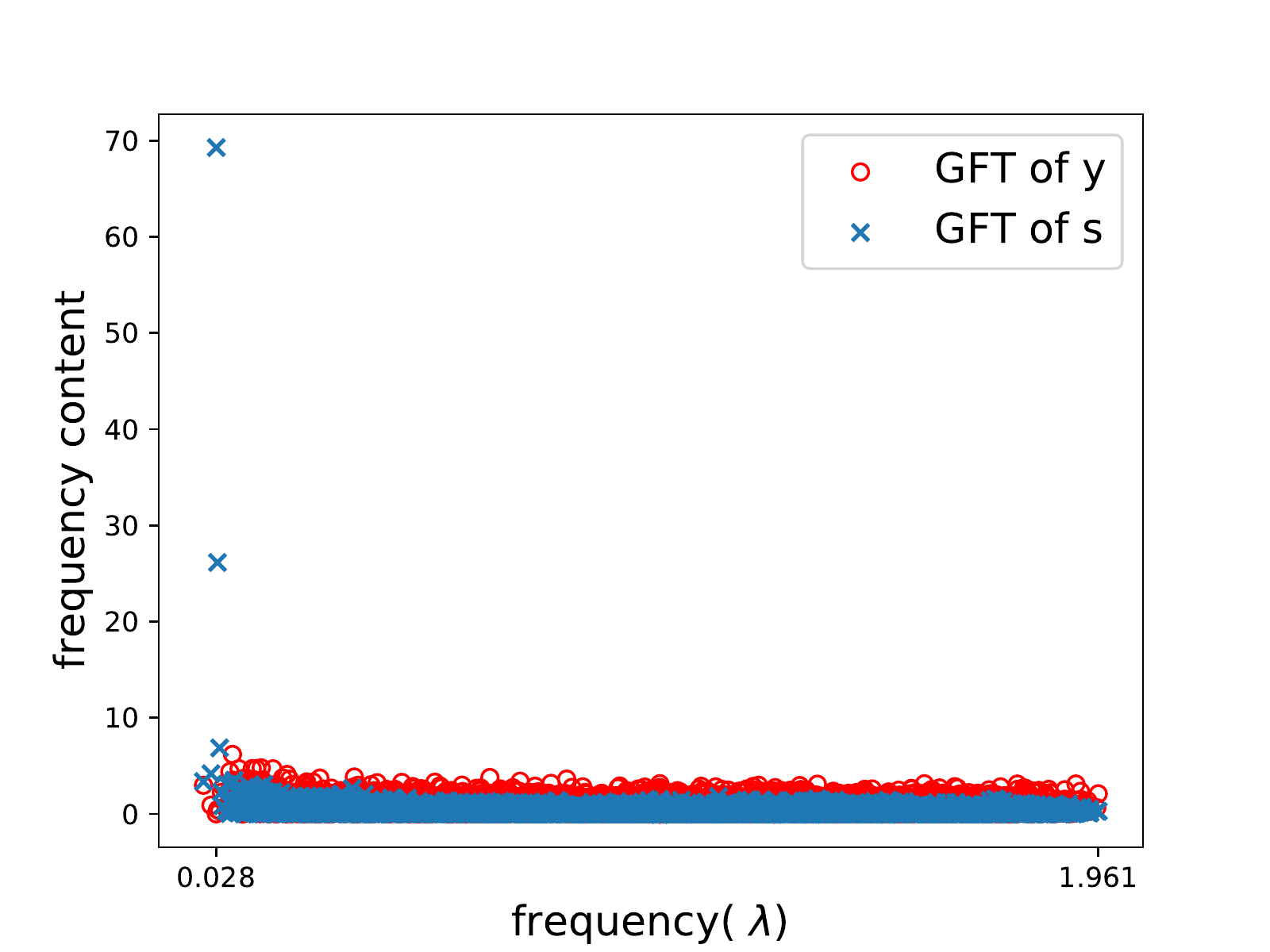}
        \caption{Pokec-z}
    \end{subfigure}
    \hskip2em
    \begin{subfigure}{0.35\textwidth}
        \includegraphics[width=\textwidth]{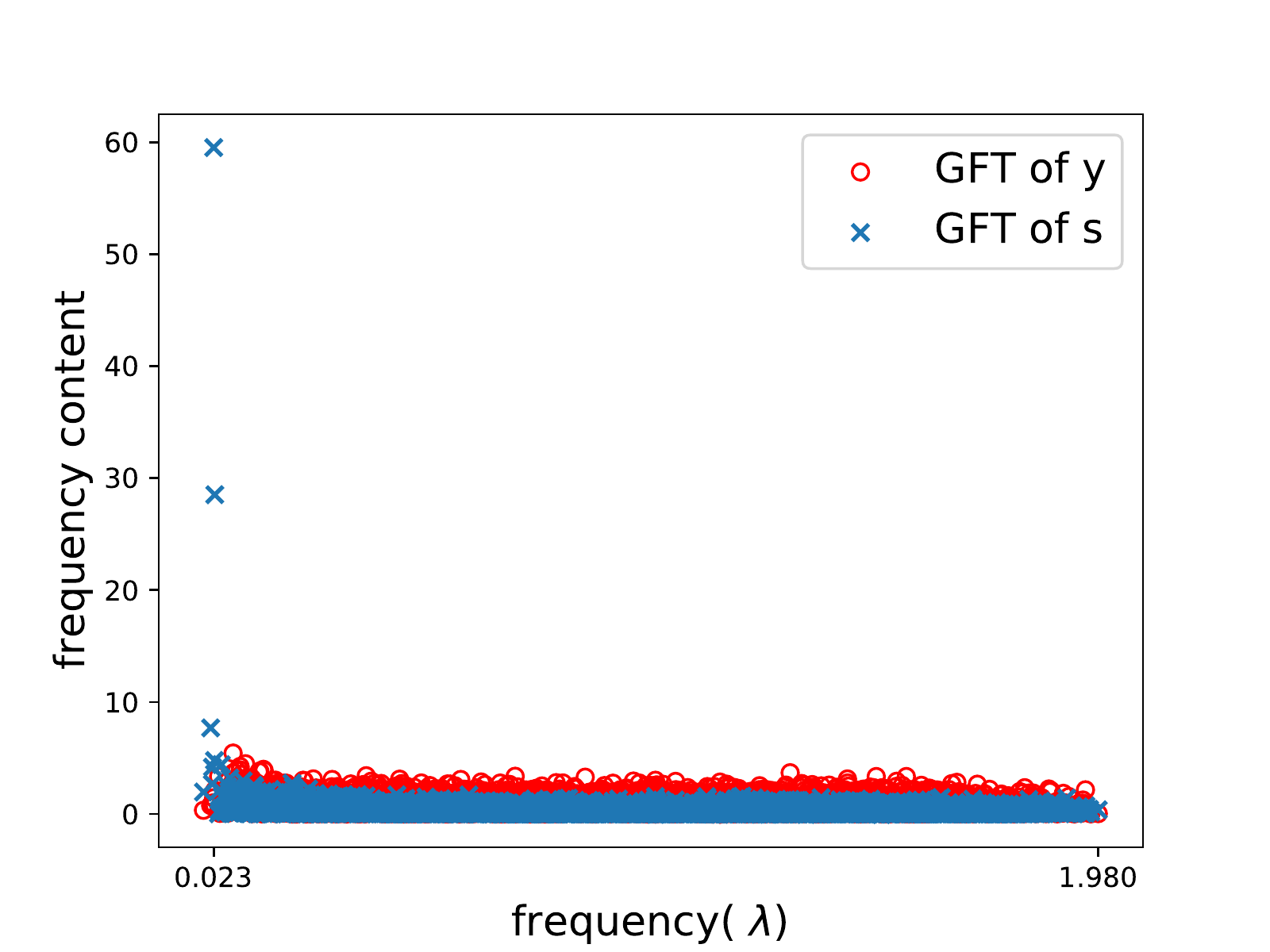}
        \caption{Pokec-n }
    \end{subfigure}
    \vspace{-2mm}
    \caption{Spectra of the graph signals $\mathbf{s}$ (sensitive attributes) and $\mathbf{y}$ (labels) over different graph frequencies. There are few low frequencies where the magnitudes of $\Tilde{\mathbf{s}}$ are markedly higher  than those of $\Tilde{\mathbf{y}}$. {\color{mpurple}, Pokec-z and Pokec-n are real-world networks, whose statistics are presented in Table \ref{table:stats}.}
    }
    \label{fig:within}
\end{figure*}

\noindent\textbf{Proposed approach and contributions.} Here, we first analyze the sources of bias inherent to the graph topology and design a general purpose bias-mitigation scheme by bringing to bear GSP notions. Our previous endeavor \cite{ouricassp} is also built upon spectral analysis of graph signals, where a fairness-aware dimensionality reduction algorithm was developed. However, in \cite{ouricassp}, the information carried in certain frequencies is completely removed, which can adversely affect the overall utility of the underlying ML task. Hence, in the present work, we design a more flexible bias mitigation algorithm that filters out the information coming from the sensitive attribute signal (e.g. race, gender in social networks), while also providing a better fairness-utility trade-off. To this end, we design a graph filter which selectively changes the graph Fourier coefficients \cite{gft} of input signals, while taking fairness into account. Overall, our contributions are:\\
\textbf{i)} Based on theoretical analysis of bias in graph topologies, a graph filter is designed to mitigate bias. The novel filter is versatile and can be employed in different stages of the learning pipeline;\\
\textbf{ii)} The proposed filter is proved to be more effective than the fairness-agnostic counterpart in  reducing bias; and\\
\textbf{iii)} Experimental results for node classification on real-world networks corroborate the effectiveness of the proposed method in mitigating bias, while providing similar utility to state-of-the-art algorithms.

\section{Preliminaries and Problem Statement}
\label{sec:methodology}
\par The focus of this study is to mitigate bias in graph-based learning algorithms by employing a graph filter for a given undirected graph $\mathcal{G}:=(\mathcal{V}, \mathcal{E})$, where $\mathcal{V}:=$ $\left\{v_{1}, v_{2}, \ldots, v_{N}\right\}$ denotes the set of nodes and $\mathcal{E} \subseteq \mathcal{V} \times \mathcal{V}$ is the set of edges. Connectivity of the input graph is encoded in the graph adjacency matrix $\mathbf{A} \in\{0,1\}^{N \times N}$, where $A_{i j}=1$ if and only if $\left(v_{i}, v_{j}\right) \in \mathcal{E}$. In addition, $\mathbf{X} \in \mathbb{R}^{N \times F}$ represents the nodal features of the input graph. The diagonal graph degree matrix is $\mathbf{D} \in \mathbb{R}^{N \times N}$, where $D_{ii}$ denotes the the degree of $v_{i}$, and $\mathbf{L} = \mathbf{I}_{N}-\mathbf{D}^{-\frac{1}{2}}\mathbf{A}\mathbf{D}^{-\frac{1}{2}}$ is the normalized graph Laplacian matrix. The sensitive attribute is defined to be the nodal feature on which the decisions should not be dependent on for fair decision making. Herein, the sensitive attribute is assumed to be binary and is denoted by $\mathbf{s} \in \{-1,1\}^{N }$. The feature vector and the sensitive attribute of node $v_{i}$ are denoted by $\mathbf{x}_{i} \in \mathbb{R}^{F}$ and $s_{i} \in \{-1,1\}$, respectively.  

The graph Fourier transform (GFT) is an orthonormal transform that provides the representation of a graph signal $\mathbf{z} \in \mathbb{R}^N$ in the graph spectral domain \cite{gft, gft2, gft3}. Specifically, taking the GFT of a graph signal amounts to projecting the signal onto a space spanned by the orthogonal eigenvectors of the positive semi-definite (PSD) normalized graph Laplacian matrix $\mathbf{L}$ \cite{gft}. Let the eigendecomposition of the normalized Laplacian be $\mathbf{L}=\mathbf{V} \mathbf{\Lambda} \mathbf{V}^{\top},$ where $\mathbf{\Lambda}=\textrm{diag}(\lambda_1,\ldots,\lambda_{N})$ collects the non-negative eigenvalues and $\mathbf{V}$ is the matrix of Laplacian eigenvectors. 
Then, the GFT of the graph signal $\mathbf{z} \in \mathbb{R}^{N}$ is given by $\Tilde{\mathbf{z}}=\mathbf{V}^{\top} \mathbf{z}$. Graph frequencies correspond to the eigenvalues of the Laplacian (a measure of smoothness of the eigenvectors with respect to the graph), meaning that the GFT decomposes signals into frequency modes (i.e., the eigenvectors of $\mathbf{L}$) of different variability over $\mathcal{G}$.

In classical signal processing, filters are utilized to manipulate signals such that their, e.g., unwanted components are attenuated or removed. Similarly, graph filters can be used to modify graph signals for different purposes including graph signal classification \cite{zhu2003semi, belkin2004semi}, smoothing and denoising \cite{zhang2008graph, shuman2011chebyshev}. Filtering an input graph signal $\mathbf{z}_{\text{in}} \in \mathbb{R}^N$ via a filter with frequency response $\tilde{\mathbf{h}}:=[\tilde{h}_{1}, \ldots,  \tilde{h}_{N}]^\top$ can be mathematically expressed as (e.g.,~\cite{gsp,gft,gama2020graphs})
\begin{equation}
\begin{split}
    \mathbf{z}_{\text{out}}&= \mathbf{V} \underbrace{\operatorname{diag}(\tilde{h}_{1}, \ldots ,\tilde{h}_{N} ) \Tilde{\mathbf{z}}_{\text{in}}}_{\text {Frequency domain filtering}}.
\end{split}
\end{equation}
Therefore, filtering in the frequency domain corresponds to point-wise multiplication of the input signal's GFT  $\Tilde{\mathbf{z}}_{\text{in}}$ with the frequency response of graph filter $\tilde{\mathbf{h}}$. In this paper, given $\mathcal{G}$ and $\mathbf{s}$, we address the problem of designing a graph filter with frequency response $\tilde{\mathbf{h}} \in \mathbb{R}^{N}$, so that the bias caused by the graph topology can be attenuated when the filter is applied to input/output graph signals in the learning algorithm.

\section{Bias Mitigating Graph Filter Design}

\subsection{Spectrum Analysis}
The homophily principle suggests that nodes with similar attributes are more likely to connect in networks, which hints at denser connectivity between the nodes with the same sensitive attributes and also with the same label \cite{segregation}. Accordingly, both the sensitive attribute $\mathbf{s}$ and node labels $\mathbf{y}$ are expected to be smooth signals over $\mathcal{G}$. This implies higher energy concentration for $\Tilde{\mathbf{s}}$ and $\Tilde{\mathbf{y}}$ over lower frequencies. However, the extent of the overlap between the spectra of  $\Tilde{\mathbf{s}}$ and $\Tilde{\mathbf{y}}$ plays an important role in fairness-aware filter design, since we want to preserve the necessary information for a downstream task (node classification in this paper) after ``filtering out'' traces of the sensitive attribute. Indeed, the spectra of $\Tilde{\mathbf{s}}$ and $\Tilde{\mathbf{y}}$ should not match completely for the feasibility of our main idea. 

To examine this, the GFT coefficients in $\Tilde{\mathbf{s}}$ and $\Tilde{\mathbf{y}}$ over different frequencies are depicted for two real-world networks in Figure \ref{fig:within}. 
Notice how the spectra of $\Tilde{\mathbf{s}}$ and $\Tilde{\mathbf{y}}$ exhibit similar characteristics. However, there are certain frequencies where the  magnitudes of $\Tilde{\mathbf{s}}$  take significantly higher values than those of $\Tilde{\mathbf{y}}$. Such observation inspires us to design a graph filter which attenuates the sensitive information while preserving the data structure necessary for downstream ML tasks.

\subsection{Bias Analysis}
Features which are correlated with the sensitive attribute lead to intrinsic bias, even when the sensitive attribute is not utilized in learning \cite{indirect}. The correlation between the input to a learning algorithm and sensitive attributes is thus a measure of the resulting bias. Motivated by this, the linear correlation between the sensitive attribute signal $\bbs$ and graph topology $\hat{\mathbf{A}}=\mathbf{D}^{-\frac{1}{2}}\mathbf{A}\mathbf{D}^{-\frac{1}{2}}$ is considered for the ensuing bias analysis. 

Several graph-based learning approaches rely on node representations obtained via local aggregration of information, a process that can be summarized as
\begin{equation}
    \mathbf{R}=\hat{\mathbf{A}}\mathbf{X}\mathbf{W},
\end{equation}
where $\mathbf{R}$ denotes the obtained  node representations, $\mathbf{X}$ is the input graph signal, and $\mathbf{W}$ represents the learnable weight matrix; see e.g.,~\cite{chami2022machine, gcn}. Hence, if a filtered graph signal $\mathbf{X}^{f}= \mathbf{V} \text{diag}(\tilde{\mathbf{h}}) \mathbf{V}^{\top} \mathbf{X}$ is input, the obtained representation becomes
\begin{equation}
    \begin{split}
        \mathbf{R}^{f}&=\hat{\mathbf{A}}\mathbf{X}^{f}\mathbf{W}\\
        &=\mathbf{V} (\mathbf{I}_{N} - \boldsymbol{\Lambda})\mathbf{V}^{\top}\mathbf{X}^{f}\mathbf{W}\\
       &=\mathbf{V} (\mathbf{I}_{N} - \boldsymbol{\Lambda} )\mathbf{V}^{\top} \mathbf{V} \text{diag}(\tilde{\mathbf{h}}) \mathbf{V}^{\top} \mathbf{X}\mathbf{W}\\
        &=\mathbf{V} (\mathbf{I}_{N} - \boldsymbol{\Lambda} )\text{diag}(\tilde{\mathbf{h}}) \mathbf{V}^{\top} \mathbf{X}\mathbf{W}\\
        &=\mathbf{A}^{f} \mathbf{X}\mathbf{W},
    \end{split}
\end{equation}
where $\mathbf{A}^{f}:=\mathbf{V} (\mathbf{I}_{N} - \boldsymbol{\Lambda} )\text{diag}(\tilde{\mathbf{h}}) \mathbf{V}^{\top}$. 
Therefore, if we feed the aggregation process with a filtered signal $\mathbf{X}^{f}$, the effective graph topology that is utilized in the information aggregation becomes $\mathbf{A}^{f}$. Building on this key observation, the linear correlation between the sensitive attributes $\mathbf{s}$ and $\mathbf{A}^f$ is employed as a bias measure, which is proportional to $\mathbf{s}^{\top} \mathbf{A}^{f}_{:,i}$ for the $i$th column of $\mathbf{A}^f$. The following Proposition reveals the sources of bias and provides an upper bound on the total correlation between $\mathbf{s}$ and $\mathbf{A}^f$, where total correlation is given by $\rho :=\|\mathbf{s}^{\top} \mathbf{A}^{f}\|_{1}$. The proof is omitted due to lack of space.
\begin{proposition}
\label{prop:corr1}
 For a filtered input graph signal $\mathbf{X}^{f}$ by a graph filter with frequency response $\tilde{\mathbf{h}}$, $\rho :=\|\mathbf{s}^{\top} \mathbf{A}^{f}\|_{1}$ can be bounded by
\begin{equation}\label{rho}
  \rho \leq \sqrt{N} \sum_{i=1}^{N} |\tilde{s}_i| |(1-\lambda_{i})| |\tilde{h}_{i}|.
\end{equation}
\end{proposition}
%
Proposition \ref{prop:corr1} shows that the linear correlation between the effective graph topology and sensitive attributes is a function of $\sum_{i=1}^{N} |\tilde{s}_i| |(1-\lambda_{i})| |\tilde{h}_{i}|$. In the sequel we design a ``matched" graph filter to reduce this term and hence the bias.  

\subsection{Fair Filter Design}
Here we design a fair filter $\tilde{\mathbf{h}}^{\text{fair}}$ to ``filter-out" the sensitive information from the bias-amplifying graph connectivity. The novel filter can be applied to input/output graph signals used in general purpose learning algorithms; see also Remark \ref{remark1}. 

From  Proposition \ref{prop:corr1}, it follows that the intrinsic bias is bounded above by $\sum_{i=1}^{N} |\tilde{s}_i| |(1-\lambda_{i})| |\tilde{h}_{i}|$. Hence, a filter can be designed that suppresses frequencies where $m_{i} := |\tilde{s}_i| |(1-\lambda_{i})|$ are largest.
%
%
To this end, let $\mathcal{C}:= \{i\:|\:m_i>\tau {m}_{\max}\}$, where $\tau$ is a hyperparameter and $m_{\max} = \max_i\{m_i\}$. {The frequency response  corresponding to larger $m_i$ are designed to reduce the value of $|\tilde{s}_i| |(1-\lambda_{i})| |\tilde{h}_{i}|=m_i |\tilde{h}_{i}|$ in the upper bound of $\rho$ in (\ref{rho}).} Accordingly, the frequency response of the fairness-aware graph filter $\tilde{\mathbf{h}}^{\text{fair}}$ is given by
\begin{align}
\label{eq:filter}
\tilde{h}_i^{\text{fair}}=\begin{cases}
    \frac{\frac{1}{N-k}\sum_{1\leq j\leq N, j\not\in \mathcal{C}} m_{j}}{m_{i}},& \text{if } i\in \mathcal{C}\\
    1,              & \text{otherwise}
\end{cases},
\end{align}
where $k$ is the cardinality of set $\mathcal{C}$. It can be observed that with the designed filter the resulting term $m_i |\tilde{h}^{\rm fair}_{i}|=\frac{1}{N-k}\sum_{1\leq j\leq N, j\not\in \mathcal{C}} m_{j} \leq m_i,  \forall i\in \mathcal{C}$. While for $i\not\in\mathcal{C}$, $m_i |\tilde{h}^{\rm fair}_{i}|=m_i$ is unchanged for frequencies that are less relevant to the bias in order to preserve the information in the original graph structure.

It is important to emphasize that $\rho$ can be minimized by setting $\tilde{\mathbf{h}}^{\text{fair}}=\mathbf{0}$, which is equivalent to filtering out all information. However, such trivial design is uninteresting, since it maximally sacrifices the utility in the ensuing task. Therefore, there needs to be a trade-off between the utility and fairness. This trade-off can be empirically adjusted via the design parameter $\tau$. Furthermore, we also theoretically demonstrate in Proposition \ref{prop:corr} that  the proposed fair filter $\tilde{\mathbf{h}}^{\text{fair}}$ decreases the upper bound on $\rho$ more effectively than a uniform fairness-agnostic filter $\tilde{\mathbf{h}}^{u}$ incurring the same amount of information loss, i.e., $\|\tilde{\mathbf{h}}^{\text{fair}}\|_1 =\|\tilde{\mathbf{h}}^{u}\|_1$.
\vspace{-0.1cm}

\begin{proposition}
\label{prop:corr}
The proposed fair filter $\tilde{\mathbf{h}}^{\textrm{fair}}$ results in a lower upper bound for the correlation measure $\rho$ between $\mathbf{s}$ and $\mathbf{A}^{f}$, when compared to the fairness-agnostic counterpart $\tilde{h}^{u}_{j}= \frac{1}{N} \sum_{i=1}^{N} \tilde{h}^{fair}_i, \forall j=1, \dots, N$, meaning
\begin{align}
  \sum_{i=1}^{N} |\tilde{s}_i| |(1-\lambda_{i})| |\tilde{h}^{\textrm{fair}}_{i}| \leq \sum_{i=1}^{N} |\tilde{s}_i| |(1-\lambda_{i})| |\tilde{h}^{u}_{i}|.
\end{align}
\end{proposition}

\begin{table*}[ht]
	\centering
\caption{Proposed filter as fairness-aware pre-processing operator for a GNN model.}

\label{table:gnn}
\begin{scriptsize}
\begin{tabular}{l c c c c c c}
\toprule
                                                    & \multicolumn{3}{{c}}{Pokec-z}  & \multicolumn{3}{{c}}{Pokec-n}                                 \\ 
\cmidrule(r){2-4} \cmidrule(r){5-7}
                       & Accuracy ($\%$) & $\Delta_{S P}$ ($\%$) & $\Delta_{E O}$ ($\%$) & Accuracy ($\%$) & $\Delta_{S P}$ ($\%$) & $\Delta_{E O}$ ($\%$) \\\midrule
{GNN} 
                   & $ \textbf{66.52} \pm 0.27$ & $6.79 \pm 2.45$  & $7.26 \pm 3.29$ & $ \mathbf{64.96} \pm 0.19$ & $6.79 \pm 2.45$  & $7.26 \pm 3.29$ 

\\ \cmidrule(r){1-7}  
{Adversarial} 
                   & $  64.26 \pm 1.79$ & $4.85 \pm 2.16$  & $5.99 \pm 2.71$ & $  64.22 \pm 0.71$ & $4.34 \pm 3.87$  & $3.84 \pm 2.71$   
    \\ \cmidrule(r){1-7}  
{\color{mpurple}EDITS \cite{dong2022edits}} 
                   & $ {\color{mpurple} 62.67 \pm 2.64}$ & ${\color{mpurple}3.17 \pm 2.49}$  & ${\color{mpurple}4.54 \pm 2.99}$ & $ {\color{mpurple} 62.67 \pm 0.51}$ & ${\color{mpurple}4.40 \pm 2.41}$  & ${\color{mpurple}5.38 \pm 1.92}$   
    \\ \cmidrule(r){1-7}  
{\color{mpurple}FairDrop \cite{spinelli2021fairdrop}} 
                   & $ {\color{mpurple} 64.36 \pm 0.95}$ & ${\color{mpurple}6.17 \pm 1.44}$  & ${\color{mpurple}6.75 \pm 0.56}$ & $ {\color{mpurple} 63.61 \pm 0.79}$ & ${\color{mpurple}\mathbf{2.75} \pm 2.20}$  & ${\color{mpurple}\mathbf{2.86} \pm 1.69}$   
    \\ \cmidrule(r){1-7}  
{$\tilde{\mathbf{h}}^{\text{fair}} + $ GNN } 
                   & $  66.29 \pm 0.27$ & $\mathbf{1.24} \pm 1.32$  & $\mathbf{2.39} \pm 1.73$ & $  64.88 \pm 0.22$ & $\mathbf{2.73} \pm 0.90$  & $3.20 \pm 1.09$  \\ 
\bottomrule
\end{tabular}
\end{scriptsize}
\end{table*}
\remark{
\label{remark1}The designed fair filter $\tilde{\mathbf{h}}^{\text{fair}}$ can be employed in a flexible way to mitigate bias for different graph-based learning algorithms. It can be applied to the graph signals that are input to, or, output from the learning algorithms. Models that are designed for attributed graphs generally utilize the information coming from both the nodal features and graph topology. Thus, the proposed filter $\tilde{\mathbf{h}}^{\text{fair}}$ can be applied to the nodal features before they are fed to the learning pipeline, in order to prevent the amplification of bias due to the graph connectivity. Alternatively, for any algorithm that outputs a graph signal (e.g., node labels in node classification), $\tilde{\mathbf{h}}^{\text{fair}}$ can be employed on the output graph signal as a fairness-aware post-processing operation. Overall, impact of the proposed fair filter can permeate several GNN-based learning frameworks in a versatile manner. }
\section{Experimental Results}
\subsection{Dataset and Experimental Setup}
\begin{table}[h]
	\centering
\caption{Dataset statistics. }
\label{table:stats}
\begin{tabular}{c c c c c c c}
\toprule
                                                    Dataset &  $|\mathcal{S}_{-1}|$ & $|\mathcal{S}_{1}|$ & Inter-edges & Intra-edges & $F$                   \\ 
\midrule
Pokec-z  & $4851$ & $2808$ & $1730$ & $39370$ & $59$ \\
Pokec-n  & $4040$ & $2145$ & $1422$ & $29220$ & $59$ \\
\bottomrule
\end{tabular}

\end{table}



\noindent\textbf{Datasets.} The performance of the proposed fair filter design is evaluated on the node classification task over real-world social networks Pokec-z and Pokec-n \cite{fairgnn}. Pokec-z and Pokec-n are the sampled versions of the 2012 Pokec network \cite{pokec}, which is a Facebook-like social network in Slovakia \cite{fairgnn}. The region of the users is utilized as the sensitive attributes, where the users are from two major regions. Labels for the node classification task are the binarized working field of the users. 
Statistical information for the utilized datasets are presented in Table \ref{table:stats}, where $\mathcal{S}_{i}$ represents the set of nodes with sensitive attribute $i$ and inter-edges connect different sensitive attributes, while intra-edges are the edges between the same sensitive attribute. Note that $N=|\mathcal{S}_{-1}|+|\mathcal{S}_{1}|$.

\noindent \textbf{Evaluation metrics.} For the utility metric of node classification, accuracy is utilized. For fairness assessment, two quantitative measures of group fairness metrics are reported, namely \textit{statistical parity}: $\Delta_{S P}=|P(\hat{y}=1 \mid s=-1)-P(\hat{y}=1 \mid s=1)|$ and \textit{equal opportunity}: $\Delta_{E O}=|P(\hat{y}=1 \mid y=1, s=-1)-P(\hat{y}=1 \mid y=1, s=1)|$,
where $y$ represents the ground truth label, and $\hat{y}$ is the predicted label. Lower values for $\Delta_{S P}$ and $\Delta_{E O}$ indicate better fairness performance \cite{fairgnn} and are more desirable.

\noindent \textbf{Implementation details. }  
\label{subsec:implementation}
In the experiments, the designed filter is employed as a pre-processing operator on the features that are input to GNN layers in a two-layer graph convolutional network \cite{gcn}. The model is trained over $40\%$ of the nodes, while the remaining nodes are equally divided to  validation and test sets. 
The hyperparameter $\tau$ is selected as $0.05$ for Pokec-z and $0.04$ for Pokec-n via grid search among the values $\{0.04, 0.05, 0.06\}$. Furthermore, adversarial regularization \cite{fairgnn}{\color{mpurple} EDITS \cite{dong2022edits}, and FairDrop \cite{spinelli2021fairdrop} are employed as fairness-aware baselines in the experiments. For adversarial regularization, the multiplier of the regularizer is tuned via a grid search among the values $\{ 0.1, 1, 10, 100, 1000\}$ (the multiplier of classification loss is assigned to be $1$). Furthermore, for EDITS, the threshold proportion is tuned among the values $\{ 0.015, 0.02, 0.29\}$, where these values are the optimized thresholds for other datasets used in the corresponding work. Finally, $\delta$ in FairDrop algorithm is tuned among the values $\{ 0.7, 0.8, 0.9\}$}. For all experiments, results are obtained for five random data splits, and their average along with the standard deviations are reported. 

\subsection{Results}
Results for GNN-based node classification are presented in Table \ref{table:gnn}. For the proposed scheme, the natural baseline is to employ the exactly same GNN model without the fair filter as a pre-processing operator (denoted as GNN in Table \ref{table:gnn}). {\color{mpurple}Moreover, Adversarial, EDITS, and FairDrop in Table \ref{table:gnn} stand for the employment of adversarial regularization in training \cite{fairgnn}, and state-of-the-art fairness-aware baselines EDITS \cite{dong2022edits}, and FairDrop \cite{spinelli2021fairdrop}, respectively. Results in Table \ref{table:gnn} demonstrate that the proposed fair filter consistently achieves better fairness measures compared to the fairness-aware baselines Adversarial and EDITS \cite{dong2022edits}, along with better utility values. While our approach outperforms FairDrop \cite{spinelli2021fairdrop} in terms of both fairness and utility on Pokec-z, FairDrop \cite{spinelli2021fairdrop} achieves lower $\Delta_{EO}$ on Pokec-n. However, this better $\Delta_{EO}$ result of FairDrop on Pokec-n is accompanied by a drop in the utility.} Furthermore, it can be observed that the employment of the proposed filter leads to the lowest standard deviation values, and therefore enhances the stability of the results. Overall, the results corroborate the efficacy of the proposed filter design in mitigating bias while also providing similar utility measures {\color{mpurple} compared to the state-of-the-art fairness-aware baselines}.



\section{Conclusion}
We developed a fairness-aware graph filter that can be flexibly employed in various graph-based ML and signal processing algorithms. Theoretical analysis on the source of bias is provided, which guides the design of the fairness-aware filter. The novel filter is provably more effective in terms of mitigating bias with the same amount of filtered information compared to the fairness-agnostic counterpart. Node classification experiments on real-world networks demonstrate that the proposed fair filter consistently provides better fairness and robustness together with similar utility compared to  baselines. {This work opens up exciting future directions. For instance, the proposed design requires a Laplacian eigendecomposition, which incurs a complexity of $\mathcal{O}(N^3)$ and may not be ideal for large-scale graphs. Computationally-efficient designs are part of our ongoing research agenda.}
\ninept
\bibliographystyle{IEEEtran}
\bibliography{main}

\end{document}